\documentclass[a4paper]{article}
\usepackage{wrapfig,floatflt,graphics}
\usepackage{fancyvrb, adjustbox}
\usepackage{amsmath}
\DeclareMathOperator*{\argmax}{argmax} 
\usepackage{INTERSPEECH2019}

\title{HyST: A Hybrid Approach for Flexible and Accurate Dialogue State Tracking}
\name{Rahul Goel, Shachi Paul, Dilek Hakkani-T\"ur}
\address{
  Amazon}
\email{{\tt \{goerahul, shachp, hakkanit\}@amazon.com}}

\begin{document}

\maketitle
\begin{abstract}
  Recent works on end-to-end trainable neural network based approaches have
  demonstrated state-of-the-art results on dialogue state tracking.  The best
  performing approaches estimate a probability distribution over all possible
  slot values. However, these approaches do not scale for large value sets
  commonly present in real-life applications and are not ideal for tracking
  slot values that were not observed in the training set. To tackle these
  issues, candidate-generation-based approaches have been proposed. These
  approaches estimate a set of values that are possible at each turn based on
  the conversation history and/or language understanding outputs, and hence
  enable state tracking over unseen values and large value sets however, they
  fall short in terms of performance in comparison to the first group. In this
  work, we analyze the performance of these two alternative dialogue state
  tracking methods, and present a hybrid approach (HyST) which learns the
  appropriate method for each slot type. To demonstrate the effectiveness of HyST on
  a rich-set of slot types, we experiment with the recently released
  MultiWOZ-2.0 multi-domain, task-oriented dialogue-dataset. Our experiments
  show that HyST scales to multi-domain applications.  Our best performing
  model results in a relative improvement of 24\% and 10\% over the previous
  SOTA and our best baseline respectively.
  %% outperforms both methods as well as previous state-of-the-art and scales to
  %% multi-domain applications.

\end{abstract}

\section{Introduction}
\label{sec:intro}
% What is dialogue state tracking
Task-oriented dialogue systems aim to enable users to  accomplish tasks
through spoken interactions.  Dialogue state tracking in task-oriented
dialogue systems has been proposed as a part of dialogue management
and aims to estimate the belief of the dialogue system on the state of
a conversation given the entire previous conversation
context~\cite{Young:2002}. In the past decade, dialogue state tracking
challenges (DSTC)~\cite{DSTC} provided datasets and a framework for
comparing a variety of methods.

In DSTC-2~\cite{DSTC2}, many systems that rely on delexicalization,
where slot values from a semantic dictionary are replaced by slot
labels, outperformed systems that rely on spoken language
understanding outputs. More recently, an end-to-end approach that
directly estimates the states from natural language input using
hierarchical recurrent neural networks (RNNs) with LSTM cells has
achieved the state-of-the art results~\cite{Bing:E2E:2017}.  However,
these approaches do not scale to real applications, where one can
observe  natural language utterances that can include previously
unseen slot value mentions and a large, possibly unlimited space of
dialogue states. To deal with this scaling issue,
\cite{mrkvsic2016neural} proposed the neural belief tracker approach
that also eliminates the need for language understanding by directly
operating on the user utterance and integrating pre-trained word
embeddings to deal with the richness in natural language.  However,
this joint approach does not scale to a large dialogue state space
as it iterates over all slot value pairs in the ontology to make a
decision.  More recently, Rastogi et al \cite{rastogi2017scalable} proposed an open
vocabulary candidate-set ranking approach, where the set of candidates
are generated from a language understanding system's hypotheses to deal
with the scalability issue.  However, this approach does not consider
multi-valued slots due to the softmax layer over all the values. Other
work relied on all possible n-grams from the conversation context as
possible values for a candidate-set and estimated probabilities for
multiple possible values~\cite{goel2018flexible}. While these methods
were shown to handle previously unseen values, their performance is
lower in comparison to the previous approaches. There are multiple
possible explanations for this. For example, the first group of
generative methods can deal with values that were not observed in user
utterances by learning to make inferences, i.e., ``fancy restaurant''
could map to the backend value of ``expensive'' for the price range
slot, whereas for the second group, the candidate-set may fail to
capture ``expensive'' in the candidate-set, as it was not observed in
the conversation context explicitly. Furthermore, some slot types may
naturally resolve to few slot values, such as days of a week, which
may have many instances observed in the training set. The first group
of generative approaches may be more appropriate for tracking such
slots.

In this paper, we analyze both the hierarchical RNN-based and the
open-vocabulary candidate-generation approaches and propose hybrid
state tracking, HyST, a hybrid approach for flexible and accurate
dialogue state tracking, which aims to learn what method to rely on
for each slot type. To investigate the appropriateness of HyST for a
rich set of domains, we experiment with the recently released
MultiWOZ-2.0 corpus~\cite{multiwoz} which includes single as well as
multi-domain interactions. These conversations include task completion
across multiple domains and allow for transfer of values between slots
of different domains, as demonstrated with an example
hotel-reservation and taxi-booking dialogue in
Table~\ref{table:conv}. When tracking dialogue state over the 7
domains included in this corpus, our baselines outperform the previous
benchmark for joint-goal accuracy (which requires estimating the
correct values for all slots of all the 7 domains). Our best hybrid approach
achieves a joint-goal accuracy of 44.22\%, which is 4.1\% (absolute) 
higher than our best baseline, resulting in an 24\% relative
improvement over the previous SOTA. 

% don't forget to mention multi-domain
% Brief description of the two approaches we examine

% Flow of the paper:
% why others methods are short sighted
% how to measure performance
% describe the dataset
% baselines?
% describe oracle performance
% give results on oracle dataset?

\begin{table}[] \small
\caption{\small An example dialogue with dialogue states after each turn. Agent
  Turns are followed by their dialogue acts in the brackets.}\label{table:conv}

  \begin{tabular}{ll}
    User:  & I need to book a hotel in the east that has 4 stars.                    \\
    \verb+Hotel+     & \verb+area=east, stars=4+                            \\
    Agent: & I can help you with that. What is your price range?                     \\
    & \begin{footnotesize}(Hotel-Request(Price)) \end{footnotesize} \\
    User:  & That doesn't matter if it has free wifi and parking.            \\
    \verb+Hotel+ & \verb+parking=yes, internet=yes+         \\
    & \verb+price=dontcare, stars=4, area=east+         \\
    Agent: & If you'd like something cheap, I recommend  Allenbell.               \\
    & \begin{footnotesize}(Hotel-Recommend(Price), Hotel-Recommend(Name)) \end{footnotesize} \\
    User:  & That sounds good,  I would also like a taxi to the hotel  \\
    &from cambridge \\
    \verb+Hotel+ & \verb+parking=yes, internet=yes+        \\
    & \verb+price=dontcare, area=east, stars=4+        \\
    \verb+Taxi+      &  \verb+departure=Cambridge+        \\
    &  \verb+destination=Allenbell+        \\
  \end{tabular}
\vspace{-8mm}
\end{table}

%% In the following sections, we briefly summarize the related work and
%% describe the two approaches as well as the hybrid method in detail. We
%% then present the data sets and experiments and discuss the results.
\begin{figure*}[t!]
  \vspace{-8mm}
    \centering
    \begin{minipage}{0.45\textwidth}
        \centering
        \includegraphics[width=1.3\textwidth]{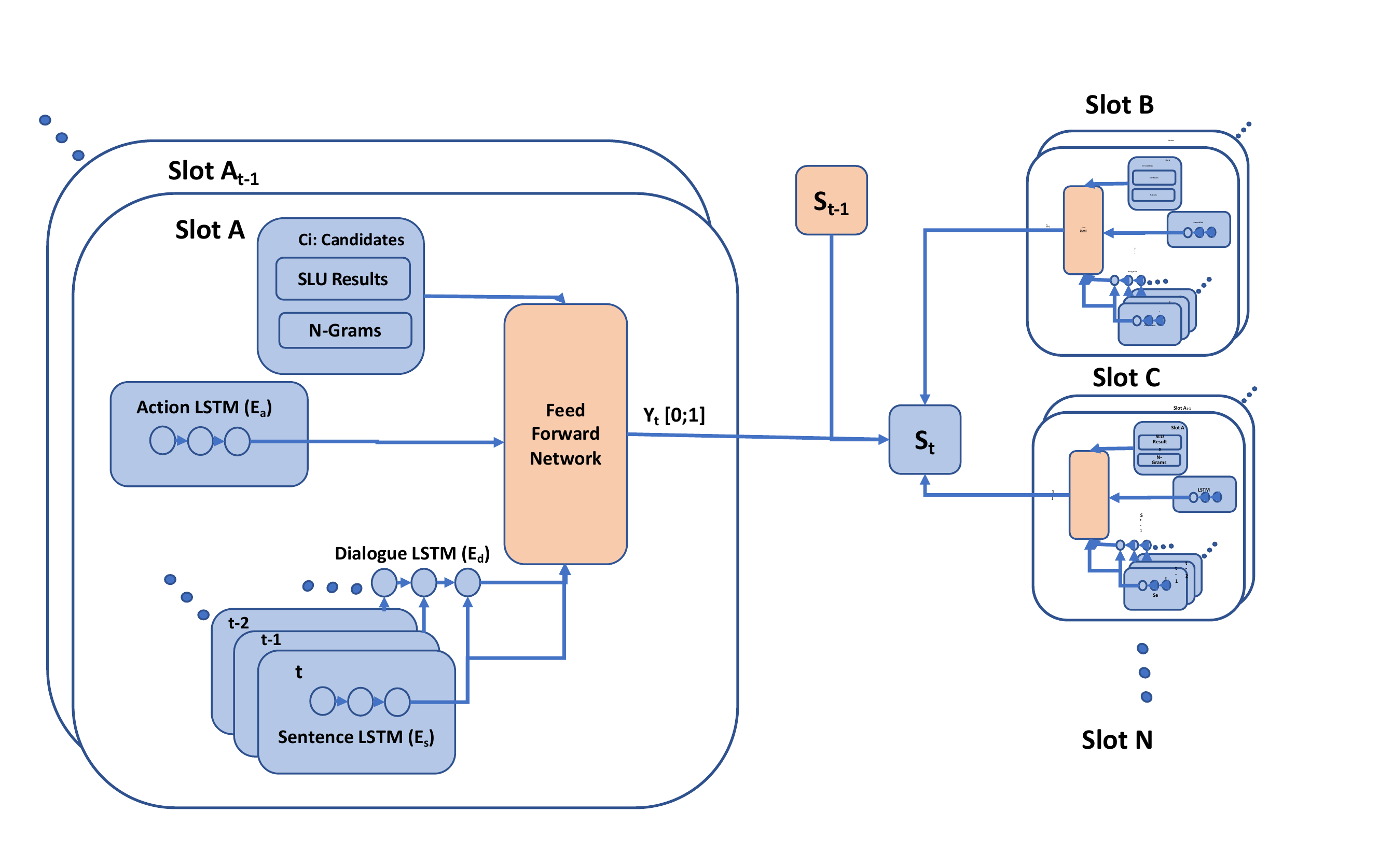} % first figure itself
        \caption{Independent model}
    \end{minipage}\hfill
    \begin{minipage}{0.45\textwidth}
        \centering
        \includegraphics[width=1.3\textwidth]{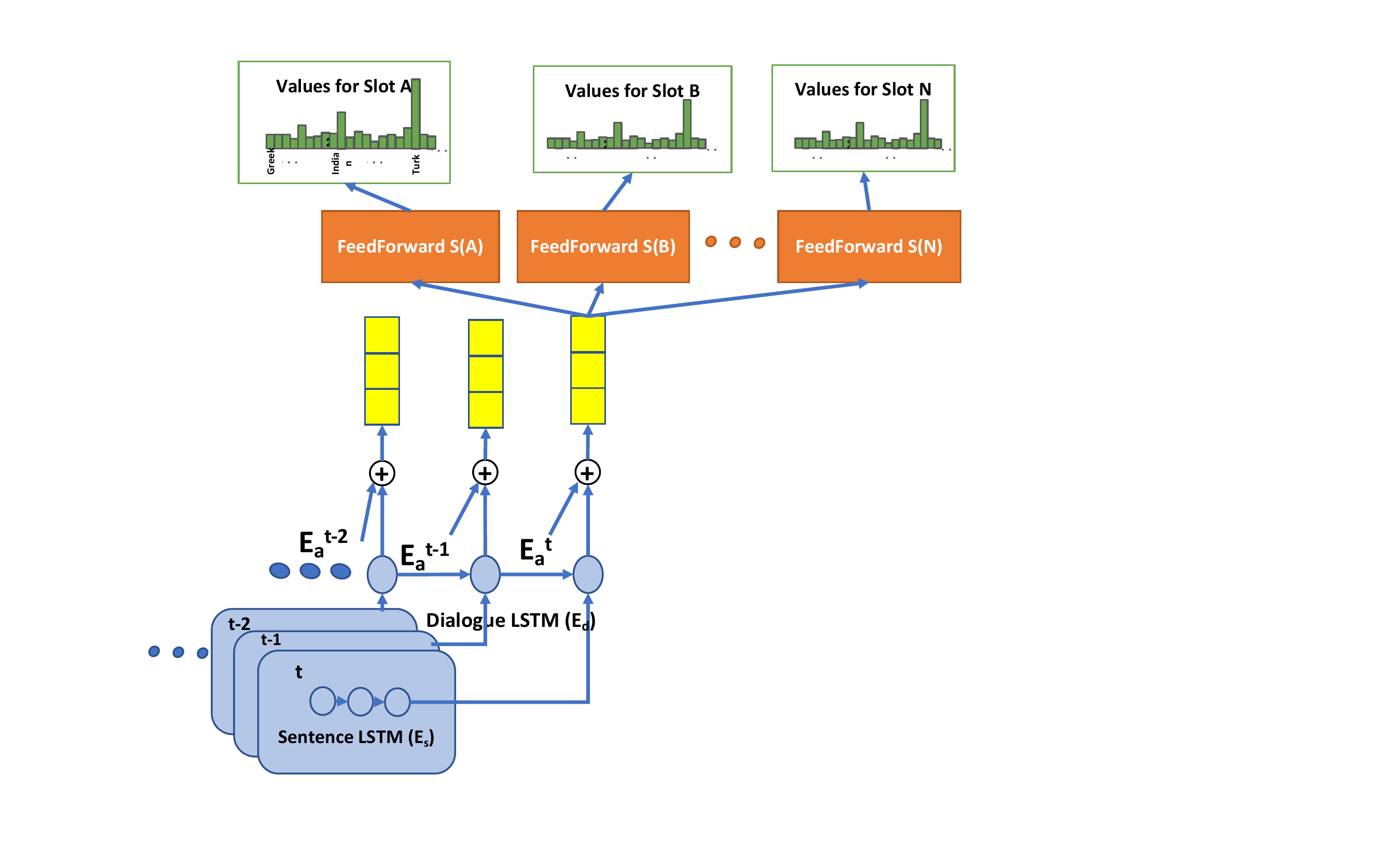} % second figure itself
        \caption{Joint model}
    \end{minipage}
    \vspace{-4mm}
\end{figure*} 

%% \begin{figure}
%%   \centering
%%       \centering
%%         \includegraphics[width=0.5\textwidth]{framework_2.pdf} % first figure itself
%%         \caption{Independent model}
%% \end{figure}
%% \begin{figure}
%%   \centering
%%   \includegraphics[width=0.5\textwidth]{framework_bing.pdf} % second figure itself
%%   \caption{Joint model}
%% \end{figure}

\section{Related Work}
\label{sec:related}

Dialogue state tracking (or belief tracking) aims to maintain a
distribution over possible dialogue states~\cite{bohus2006k,
  Williams2007partially}, which are often represented as a set of
key-value pairs. The dialogue states are then used when interacting
with the external back-end knowledge base or action sources in
determining what the next system action should be. Previous work on
dialogue state tracking include rule-based
approaches~\cite{wang2013simple}, Bayesian
networks~\cite{thomson2010bayesian}, conditional random fields
(CRF)~\cite{lee2013recipe}, recurrent neural
networks~\cite{henderson2014word}, end-to-end memory
networks~\cite{PerezLiu2017}, pointer networks~\cite{xu2018end} and
embedding-based approaches~\cite{mrkvsic2016neural,universalDST}.

Previous work that investigated joint language understanding and
dialogue state tracking include work by
\cite{Bing:E2E:2017,rastogi2018multi}.  Our hierarchical RNN approach
is inspired by \cite{Bing:E2E:2017} and uses a hierarchical recurrent
neural network to represent utterances and the dialogue flow. This
approach estimates a probability for all possible values, and hence
suffers from the scalability issues. Our hybrid approach aims to
tackle this issue by learning to switch to a candidate-generation-based approach.

For multi-domain dialogues, previous work \cite{nouri2018toward}
presented results for two approaches, global-locally self-attentive
dialogue state tracker (GLAD)~\cite{zhong2018global} and globally
conditioned encoder (GCE)~\cite{nouri2018toward}. GLAD is composed of
an encoder and scoring modules, where the encoder uses global
biLSTM modules to share parameters between
estimators for all the slots and local biLSTM modules to learn
slot-specific features. GCE is based on GLAD, but it simplifies the GLAD
encoder by removing slot-specific recurrent and self attention
layers. In our experiments, we use these two approches as baselines.

%% Will add reviews of the following in the final version:
%% \cite{nouri2018toward} and \cite{universalDST}.

\section{Methodology}\label{sec:models}

A dialogue $D$ with $N$ turns is denoted as a series of agent~($a_i$) and
user~($u_i$) turns i.e. $a_1, u_1, a_2, u_2, ..., a_N, u_N$.  The task of state
tracking is to predict the state ($S_i$) after each user turn, $u_i$, of the
conversation. The conversation state ($S_i$) is commonly defined as a set of
slot values, $s_i^k$, for slot types $s^k$, where $k \in \{1,...,T\}$ which are
predefined. We define and implement the two following prevailing approaches to
dialogue state tracking.

\subsection{Approach 1: Open Vocabulary State tracking (OV ST)}

Similar to ~\cite{goel2018flexible}, we adopt an open-vocabulary scoring model
for each slot type that needs to be tracked. The input to the model after user
turn $u_i$ is a set of candidates $\{1, \dots, |C_i|\}$, where $|C_i|$ is
  number of candidates in turn $i$ that could be a value for each slot type
  $s^k$, and the conversation context $D_i$ ($a_1, u_1, ..., a_i,u_i$).  For each
  candidate $c_i^j$ and for each slot type $s^k$, the model makes a binary
  decision $\hat y^{jk}_i \in [0,1]$, that denotes $c_i^j$ to be one of the
  values of that slot type.
% Dilek: made the index of y ijk, does it make sense to you too or is it too crowded?

If $\hat{y}^{jk}_i = 1$, we update the value of slot $s_i^k$ with
candidate $c_i^j$. For a given dialogue, we maximize the following objective:

\begin{equation}
  L(D) = \sum_{i=1}^{N}\sum_{k=1}^{T}\sum_{j=1}^{C_i}\log P\left(y^{jk}_i|c_i^j,s_i^k,D_i\right)
\end{equation}

Given a user turn $u_i$ , we construct a candidate set for that turn. A
candidate set is an open set consisting of possible slot values for each slot
type. In a typical dialogue system this could be constructed from the output of
a SLU system augmented with additional values obtained using simple rules~(such as
business logic or entity resolvers). For our experiments, we formed the % Do we need to clarify this?
candidate sets to include all word $n$-grams in user and agent utterances up-to
turn $i$ in that dialogue.  To reduce the size of the total candidate set, we
only include those n-grams that were seen as possible slot values in the
training set. We extend the candidate set with \{\verb+yes, no, dontcare+\} as
they are implied values which do not appear explicitly in the conversation.  In
practice, to increase coverage, such a system could also include additional
values like synonyms, ASR corrections, values from a knowledge graph and
resolved entities from a entity resolution system.

The system starts with a default state for every slot. After each
user utterance, we update our dialogue state with candidates that are predicted
as positive. Based on the system design, various update strategies or
constraints can be incorporated in the dialogue state update step. For example,
if we want to enforce the constraint that one slot can have only one value, we
can select the candidate with the highest score from the pool of positive
candidates. The dialogue history context features are flexible and we can easily
add new context features by appending them to the existing context vector. For
our experiments we use the following context features at each user turn $u_i$.
\begin{enumerate}
\item User utterance encoder ($E_i$):  We use a biLSTM to encode
  each utterance, $u_i=w_1^i,...,w_{n_i}^i$, where $n_i$ denotes the number of
  tokens in $u_i$ and the final utterance representation for utterance $e_i$ is
  obtained by concatenating the last hidden state of the forward lstm,
  $\overrightarrow{LSTM}$ and the first hidden state of the backward lstm,
  $\overleftarrow{LSTM}$.
  $$E_i = \overleftarrow{LSTM^{sent}}(u_i) \oplus \overrightarrow{LSTM^{sent}}(u_i)$$
  
\item Hierarchical LSTM ($Z_i$): We use a unidirectional LSTM over past user
  utterances to encode the dialogue context.
  \begin{equation}\label{hier}
    Z_i = LSTM^{dialogue}(E_1,...E_i)
  \end{equation}
  
\item Dialogue Act LSTM ($A_i$): We use a unidirectional LSTM over agent
  dialogue acts to encode agent dialogue acts.
  $$LSTM^{dialogueAct}(s_1,...s_k)$$
\end{enumerate}

We concatenate all of these features into a context feature vector
$F_{context}$. The context encoders are shared for all slots.  For every
slot type, we have:
\begin{equation} \label{context}
  F_{context} = [E_i;Z_i;A_i]
\end {equation}
\begin{equation}
  \hat{y_j} = \mbox{sigmoid}(FF_k(c_i^j, F_{context})).
\end{equation}
The final layer $FF_k$ is a feed forward layer, which estimates the probability
of $c_i^j$ filling slot $k$.

\subsection{Approach 2: Joint State tracking (JST)}
The joint state tracking approach builds a hierarchical RNN modeling words and
turns of each dialogue~\cite{Bing:E2E:2017}. Similar to the open vocabulary
state tracking approach, we obtain the dialogue representation $Z_i$
(Equation \ref{hier}).The final layer is a feed forward network for each slot
type $k$, $FF_k$, which estimates a probability distribution over all the
possible values for that slot type, $S_k = s_k^1,...,s_k^{|S_k|}$:
\begin{equation}
P_i(s_k \in S_k| Z_i) = \mbox{softmax}_k(FF_k(Z_i))
\end{equation}
The vocabulary of possible values, $V_k$, is formed of the values
observed in the training set, including \verb+none+, and \verb+dontcare+. The
hierarchical RNN layers are shared for all the slot types.

\subsection{Hybrid state tracking (HyST)}
We combine the two aforementioned approaches into a hybrid approach. For each
slot we choose between OV ST and JST.  Let $$A^k(M) = \frac{\sum_{i=1}^{N} 1\{y_{i}^k = \hat{y}_{i}^{kM}\}}{N}$$
be the accuracy of slot $k$ over given a approach $M$.  For each slot we pick
the optimal approach $M_{opt}$ as $$ M_{opt} = \argmax_{JST, OVST}(A^k(JST),
A^k(OVST)).$$ We learn the approach to pick using our
development set. The slots on which the open vocabulary approach performed better on
the development set are marked with `*' in Table~\ref{table:slots}.

%% it uses a bi-directional LSTM
%% to encode each utterance, $u_i=w_1^i,...,w_{n_i}^i$, where $n_i$ 
%% denotes the number of tokens in $u_i$ and the final utterance
%% representation for utterance $E_i$ is obtained by concatenating the
%% last hidden layer of the forward LSTM, $\overrightarrow{LSTM}$ and the
%% first hidden layer of the backward LSTM, $\overleftarrow{LSTM}$, as
%% well as an embedding of the system action at turn i, $s_i$:
%%  $$E_i = \overleftarrow{LSTM}(u_i) \oplus \overrightarrow{LSTM}(u_i)
%% \oplus s_i $$
%% Then a hierarchical, uni-directional LSTM is used to
%% represent the dialogue level information for each time step, $Z_i$:
%% $$Z_i = LSTM(E_1,...E_i) $$

\section{Data}\label{sec:data}
\label{sec:data}
For our state tracking experiments we use the MultiWOZ-2.0
dataset~\cite{multiwoz}.The  MultiWOZ-2.0 dataset consists of multi-domain
conversations from 7 domains with a total of 37 slots across domains. Some of
the slots, for example, \verb+day+ and \verb+people+, occur in multiple domains. An example
conversation is shown in Table~\ref{table:conv}. For our experiments, we treat
each slot independently and do not share slots between domains. So, the same
slot type is present in several domains and is represented by appending domain
and slot names as \verb+domain.slot+ in Table~\ref{table:slots}.  If a user turn
does not have a slot value assigned to it, we mark it as \verb+None+. Some
dataset statistics are shown in Table~\ref{table:data}. We also present a
detailed breakdown of different slot types in MultiWOZ-2.0 in
Table~\ref{table:slots}. The table includes out-of-vocabulary (OOV) slot value
rates in the development set for each slot type. This is computed as the
percentage of values of each slot type in the development set that was not
observed in the training partition.

To showcase the complexity for different slot types, we also include percentage
of turns with a ``None'' value for each slot type and the number of unique
values (i.e., Vocabulary size) for each slot. The final row of
Table~\ref{table:slots} presents the percentage of turns whose complete state has
never been observed in the training set for JST and in the candidate set
generated by our OV ST approach.
%% According to these two numbers, the upper bound for the
%% joint-goal accuracy of the JST approach is 97.52\% and OV ST approach is
%% 71.82\%.

\begin{table}[]
\caption{Some dataset statistics. Numerical values refer to things like `time'
  and `people' which are open ended. }\label{table:data}
\scalebox{0.8}{
  \begin{tabular}{l|lll}
  \textbf{Data property}                    & \textbf{Train}  & \textbf{Dev} & \textbf{Test}  \\
\hline
\# Dialogues                     & 8,483  & 1,000 & 1000 \\
\# User turns                    & 56,781 & 7,374 & 7372 \\
\# Uservocab (with num. values)        & 4311   & 1875 & 1840  \\
\# Uservocab (without num. values)        & 3805   & 1709 & 1646  \\
Median user sent length (tokens) & 11     & 11  & 11
\end{tabular}}
  \vspace{-4mm}%
\end{table}

%% \dbltextfloatsep{-5mm}
%% \dblfloatsep{-5mm} 
%\vspace{-4em}
\begin{table}[t!]
  %% \setlength\belowcaptionskip{-20pt}
  %%   \setlength\abovecaptionskip{-20pt}
    %\vspace{-15mm}
\centering
  \caption{\small {Slot breakdown for MultiWOZ-2.0. The vocab size is the number
      of unique values seen for a slot in the training set. \%None refers to
      the percentage of turns where the slot was not set. Out-of-vocabulary
      (OOV) rate refers to percentage of slots in the development set which are
      not present in the training set for each method. Slots marked with * are
      the ones which had better performance using the open-vocabulary (OV)
      approach. }}\label{table:slots}
\scalebox{0.9}{
    \begin{tabular}{l|llll}
      \textbf{Slot Name}             & \textbf{Vocab} & \textbf{\%None} & \textbf{OV} & \textbf{JST} \\ 
      &  \textbf{Size} &  & \textbf{OOV rate} & \textbf{OOV rate} \\ \hline
      taxi.leaveAt*          & 119        & 96.19\% &1.25\%  &0.19\% \\  
      taxi.destination*      & 277        & 92.76\% &0.80\%  &0.14\% \\
      taxi.departure*        & 261        & 92.87\% &0.94\%  &0.34\% \\
      taxi.arriveBy*         & 101        & 96.84\% &0.39\%  &0.08\% \\
      restaurant.people     & 9          & 84.10\%  &0.72\% &0.00\% \\
      restaurant.day        & 10         & 84.11\%  &0.18\% &0.00\% \\
      restaurant.time*       & 61         & 84.22\% &0.27\%  &0.05\% \\
      restaurant.food       & 104        & 71.65\%  &0.60\% &0.00\%\\
      restaurant.pricerange & 11         & 74.62\%  &0.58\% &0.00\%\\
      restaurant.name*       & 183        & 87.14\% &1.48\%  &0.26\%\\
      restaurant.area       & 19         & 74.02\%  &0.47\% &0.01\%\\
      bus.people            & 1          & 100\%    &0.00\% &0.00\%\\
      bus.leaveAt           & 2          & 99.99\%  &0.00\% &0.00\%\\
      bus.destination       & 5          & 99.94\%  &0.00\% &0.00\%\\
      bus.day               & 2          & 99.99\%  &0.00\% &0.00\%\\
      bus.arriveBy          & 1          & 100\%    &0.00\% &0.00\%\\
      bus.departure         & 2          & 99.94\%  &0.00\% &0.00\%\\
      hospital.department   & 52         & 99.30\%  &0.09\% &0.00\%\\
      hotel.people          & 9          & 84.61\%  &0.83\% &0.00\%\\
      hotel.day             & 11         & 84.59\%  &0.18\% &0.00\%\\
      hotel.stay            & 9          & 84.64\%  &0.66\% &0.00\%\\
      hotel.name            & 89         & 84.80\%  &1.68\% &0.24\%\\
      hotel.area            & 24         & 80.82\%  &0.18\% &0.00\%\\
      hotel.parking         & 8          & 85.58\%  &0.45\% &0.00\%\\
      hotel.pricerange      & 9          & 82.74\%  &0.72\% &0.00\%\\
      hotel.stars           & 13         & 83.59\%  &1.83\% &0.01\%\\
      hotel.internet        & 8          & 85.88\%  &0.49\% &0.00\%\\
      hotel.type            & 18         & 82.18\%  &1.49\% &0.07\%\\
      attraction.type       & 37         & 81.45\%  &3.95\% &0.01\%\\
      attraction.name*       & 137        & 89.71\% &1.37\%  &0.38\%\\
      attraction.area       & 16         & 82.80\%  &0.33\% &0.03\%\\
      train.people          & 13         & 89.11\%  &2.54\% &0.00\%\\
      train.leaveAt*         & 134        & 86.68\% &3.09\%  &0.72\%\\
      train.destination     & 29         & 71.89\%  &1.00\% &0.05\%\\
      train.day             & 11         & 72.90\%  &0.20\% &0.04\%\\
      train.arriveBy*        & 107        & 86.81\% &1.82\%  &0.18\%\\
      train.departure       & 35         & 72.38\%  &0.94\% &0.07\%\\
      All values &        &                &25.60\% & 2.48\% \\
%      Open-vocab oracle &     &      & 28.18\% \\
  \end{tabular}}
  \vspace{-4mm}%
\end{table}

\section{Experimental Setup}
\label{sec:experiments}

In all experiments, we clip each turn to 30 tokens and each dialogue to past 30
turns. We use ADAM~\cite{kingma2014adam} for optimization with a learning rate
of 0.001 and default parameters. We use a batch size of 128 while training. We
initialize our embedding matrices randomly and learn them during training. We
use manual search to tune all our parameters using our development set.

\textbf{Open vocabulary state tracking:} The model consists of four encoders: the
sentence encoder, hierarchical dialogue encoder, dialogue act encoder and the
candidate encoder. Our candidate encoder is an embedding lookup of dimension
300. We use the same embedding layer as input to the sentence encoder. Our
sentence encoder is a biLSTM with hidden size of 256. Our sentence
representation is the final state of the biLSTM. The hierarchical dialogue encoder
is a LSTM with hidden size of 512 which takes the sentence
representation as input. We use an embedding size of 50 for system dialog acts
and encode them using a LSTM with an hidden size of 64. We concatenate
these representations and pass it through a feed-forward network with an output of
256. The final 256-dimensional vector is used for a binary decision per slot
type.

\textbf{Joint state tracking:} The joint model represents words and system
actions with 300-dimensional vectors, with a hidden layer size of 200 for the
utterance LSTM and 150 for the dialogue level LSTM. The agent actions were
found to be not useful in the early experiments and are
excluded from the final results.

\section{Results}
\label{sec:results}
We present per domain results in Table~\ref{table:results}. As in
previous work, we report joint goal accuracy as our metric. For each
user turn, we get the joint goal correct if our predicted state
exactly matches the ground truth state for all the slots in that domain. As our
candidate set generation is based on n-grams OOV rate for the OV oracle
(Table~\ref{table:slots}) is high. This implies that the performance ceiling for
this approach is around 74.4\% which is much lower than the ceiling for the JST
(97.5\%). Still, we observe that for the slots with large vocabulary sizes~(ones
marked with * in Table ~\ref{table:slots}), the OV approach outperforms the
joint model. All slots with over 100 possible values with the exception of one,
restaurant.food with a vocab size of 104, were better tracked with the OV
approach. Combining the two approaches into a hybrid approach leads to the best
performance on most domains except 'Hotel' where there was no significant change.

\begin{table}[]
  \centering
  \caption{Comparison of various apporaches on different domains. The numbers
    presented are joint goal accuracy.}\label{table:results}
  \scalebox{0.9}{
  \begin{tabular}{l|lll}
    \textbf{Domain}         & \textbf{JST} & \textbf{OV ST} & \textbf{HyST} \\ \hline
    Taxi        & 91.48\%   & 92.30\%           & 92.30\%         \\
    Restaurant  & 78.55\%   & 75.05\%           & 79.67\%         \\
    Bus        & 100\%     & 99.95\%           & 100\%           \\
    Hospital    & 100\%     & 100\%             & 100\%           \\
    Hotel       & 79.38\%   & 74.04\%           & 79.14\%         \\
    Attraction & 83.42\%   & 82.80\%           & 85.63\%         \\
    Train       & 82.13\%   & 71.19\%           & 84.03\%         \\ 

  \end{tabular}}
    \vspace{-3mm}%
\end{table}

Table~\ref{table:results} presents the joint goal accuracy for each
domain with the three approaches. From this table, we observe a large
difference in the 2 approaches for domains like Train and Hotels. One
reason for this is the implicit entity resolution which is
part of the dataset. For example, when a user says: `I want a taxi to
petersbrough', the state in MultiWOZ-2.0 gets updated as `To:
Petersbrough, From: Cambridge'. The city Cambridge is never mentioned
by the user or the agent but is implicit due to task location. JST is
able to infer these relations due to similar data patterns but for OV
ST, the word `Cambridge' never occurs in the candidate set leading to
lower performance.

We present overall joint accuracy on MultiWOZ-2.0 in
Table~\ref{table:overall}. We also report results on a few baselines. The first
is a dumb baseline which assigns \verb+none+ (the majority class) to all slots. We
observe that this is a very weak baseline due to the large label space. Second,
we present the benchmark baseline from the MultiWOZ-2.0 corpus~\cite{multiwoz}
webpage\footnote{http://dialogue.mi.eng.cam.ac.uk/index.php/corpus/}.  We also
report results on GLAD~\cite{zhong2018global} and its extension
GCE~\cite{nouri2018toward}.  The hybrid approach is better than all baselines,
and results in a joint goal accuracy of 44.24\% when all the 7 domains and 37
slots are considered.

\begin{table}[]
  \centering
  \caption{Joint goal Accuracy on MultiWOZ-2.0. We present the ensemble (we do three
    independent model runs and take the majority vote per slot) results for our
    methods (presented in \textbf{boldface}) with the single model results in
    parentheses.  }\label{table:overall}
  \scalebox{0.9}{
  \begin{tabular}{l|lll}    
    \textbf{Method}         &  \textbf{Accuracy}      \\ \hline
    Majority Baseline         & 1.5\%     \\
    MultiWOZ-2.0 Benchmark & 25.83\% \\
    \textbf{OV ST}     & 31.11\% (29.73\%)      \\
    GLAD \cite{zhong2018global,nouri2018toward}  & 35.57\% \\
    Previous SOTA (GCE) \cite{nouri2018toward}  & 35.58\%     \\
    \textbf{JST}    & 40.74\% (38.42\%)       \\
    \textbf{HyST}       & 44.24\% (42.33\%)     \\
  \end{tabular}}
    \vspace{-6mm}%
\end{table}

\section{Conclusions}
\label{sec:conclusion}
The joint tracking approach couples spoken language understanding and dialogue
state tracking to achieve high accuracy on state tracking benchmarks, but this
limits its performance on slots with large vocabulary as shown in our
experiments. On the other hand the open-vocabulary approach is very flexible and
shows better performance on large-vocabulary slots.  In this work we presented
HyST, a hybrid approach for dialogue state tracking by combining the aforementioned
approaches. By learning to switch between the 2 approaches, our approach
outperforms both of them on the challenging MultiWOZ-2.0 corpus. HyST achieves
44.2\% joint goal accuracy on MultiWOZ-2.0 beating previous SOTA by over
24\% (relative). Going forward we would like to experiment with better candidate
representations for the OV ST approach. One exciting follow up would be
enabling zero-shot state tracking by copying over values which have appeared in
previous states to new domains.

\bibliographystyle{IEEEtran}

\bibliography{mybib}

\end{document}